\RequirePackage{fix-cm}
\documentclass[smallextended]{svjour3}       
\smartqed  
\usepackage{svg}
\usepackage{comment}
\usepackage{graphicx}
\usepackage{natbib}
\usepackage{ismll}
\usepackage{amsmath}
\usepackage{amssymb}
\usepackage{mathtools}
\usepackage{multirow}
\usepackage{hyperref}
\usepackage{algorithm}
\usepackage{xcolor}
\usepackage{dcolumn}
\usepackage{algorithmic}
\usepackage{subfig}
\usepackage{booktabs}
\renewcommand{\algorithmicrequire}{\textbf{Input: }}

\newcommand{\rpm}{\sbox0{$1$}\sbox2{$\scriptstyle\pm$}
  \raise\dimexpr(\ht0-\ht2)/2\relax\box2 }
\newcommand\meta{^{\text{meta}}}

\begin{document}

\title{Dataset2Vec: Learning Dataset Meta-Features
}


\author{Hadi S. Jomaa         \and
        Lars Schmidt-Thieme \and
        Josif Grabocka
}

\authorrunning{Jomaa et. al} 

\institute{Hadi S. Jomaa \at
          University of Hildesheim, Samelsonplatz 1, 31141 Hildesheim, Germany \\
          \email{hsjomaa@ismll.uni-hildesheim.de}
           \and
           Lars Schmidt-Thieme \at
          University of Hildesheim, Samelsonplatz 1, 31141 Hildesheim, Germany \\
          \email{schmidt-thieme@ismll.uni-hildesheim.de}
          \and
           Josif Grabocka\at
          University of Freiburg, Georges-Köhler-Allee 74, 79110 Freiburg, Germany \\
          \email{grabocka@informatik.uni-freiburg.de}               
}

\date{Received: date / Accepted: date}

\maketitle

\begin{abstract} \small\baselineskip=9pt
Meta-learning, or learning to learn, is a machine learning approach
that utilizes prior learning experiences to expedite the
learning process on unseen tasks. As a data-driven
approach, meta-learning requires meta-features that represent the
primary learning tasks or datasets, and are estimated traditonally as engineered dataset 
statistics that require expert domain knowledge tailored for every meta-task. 
In this paper, first, we propose a meta-feature extractor called Dataset2Vec that combines the
versatility of engineered dataset meta-features with the expressivity of meta-features learned by
deep neural networks. Primary learning tasks or datasets are represented
as hierarchical sets, i.e., as a set of sets, esp. as a set of predictor/target
pairs, and then a DeepSet architecture is employed to regress meta-features on them.
Second, we propose a novel auxiliary meta-learning task with abundant data called dataset similarity learning
that aims to predict if two batches stem from the same dataset or different ones.
In an experiment on a large-scale hyperparameter optimization task for 120 UCI datasets
with varying schemas as a meta-learning task, we show that the meta-features of Dataset2Vec
outperform the expert engineered meta-features and thus demonstrate the usefulness of
learned meta-features for datasets with varying schemas for the first time.

\end{abstract}

\section{Introduction}

Meta-learning, or learning to learn, refers to any learning approach
that systematically makes use of prior learning experiences to
accelerate training on unseen tasks or datasets~(\cite{vanschoren2018meta}).
For example, after having chosen hyperparameters for dozens of different learning
tasks, one would like to learn how to choose them for the next task at hand.
Hyperparameter optimization across different datasets is a typical meta-learning task
that has shown  great success lately~(\cite{DBLP:conf/icml/BardenetBKS13,DBLP:journals/ml/WistubaSS18,DBLP:conf/aistats/YogatamaM14}).
Domain adaptation  and learning to optimize are other such meta-tasks of interest~(\cite{DBLP:conf/icml/FinnAL17,DBLP:journals/corr/abs-1807-05960,DBLP:conf/nips/FinnXL18}).

As a data-driven approach, meta-learning requires meta-features that represent the
primary learning tasks or datasets to transfer knowledge across them.
Traditionally, simple, easy to compute,
engineered~(\cite{DBLP:journals/corr/abs-1808-10406}) meta-features,
such as
   the number of instances,
   the number of predictors,
   the number of targets~(\cite{DBLP:conf/icml/BardenetBKS13}),
   etc.,
have been employed.
More recently, unsupervised methods based on variational autoencoders~(\cite{DBLP:conf/iclr/EdwardsS17})
have been successful in \textit{learning} such meta-features.
However, both approaches suffer from complementary weaknesses.
  Engineered meta-features often require expert domain knowledge and must be adjusted for each task,
   hence have limited expressivity.
On the other hand,
   meta-feature extractors modeled as autoencoders can only compute meta-features for
   datasets having the same schema, i.e. the same number, type, and semantics of predictors and targets.

Thus to be useful, meta-feature extractors should meet the following four desiderata:
\begin{itemize}
\item[] \textbf{D1. Schema Agnosticism:} The meta-feature extractor should be able to extract meta-features for
  a population of meta-tasks with varying schema, e.g., datasets containing different
  predictor and target variables, also having a different number of predictors and
  targets.
  
\item[] \textbf{D2. Expressivity:} The meta-feature extractor should be able to extract meta-features for meta-tasks of
  varying complexity, i.e., just a handful of meta-features for simple meta-tasks, but
  hundreds of meta-features for more complex tasks.

\item[] \textbf{D3. Scalability:} The meta-feature extractor should be able to extract meta-features fast, e.g., it should not
  require itself some sort of training on new meta-tasks.

\item[] \textbf{D4. Correlation:} The meta-features extracted by the meta-feature extractor should correlate well with
  the meta-targets, i.e., improve the performance on meta-tasks such as hyperparameter optimization.
\end{itemize}

In this paper, we formalize the problem of \textbf{meta-feature learning}
as a step that can be shared between all kinds of meta-tasks and asks for
meta-feature extractors that combine the versatility of engineered meta-features
with the expressivity obtained by learned models such as neural networks,
to transfer meta-knowledge across (tabular) datasets with varying schemas (Section~\ref{section: mfl}).

First, we design a novel \textbf{meta-feature extractor} called \textbf{Dataset2Vec},
that learns meta-features from (tabular) datasets of a varying number of
instances, predictors, or targets. Dataset2Vec makes use of
representing primary learning tasks or datasets as hierarchical sets,
i.e., as a set of sets, specifically as a set of predictor/target pairs, and then
uses a DeepSet architecture~(\cite{DBLP:conf/nips/ZaheerKRPSS17})
to regress meta-features on them (Section~\ref{section: d2v}).

As meta-tasks often have only a limited size of some hundred or thousand
observations, it turns out to be difficult to learn an expressive
meta-feature extractor solely end-to-end on a single meta-task
at hand. We, therefore, second, propose a novel \textbf{meta-task}
called \textbf{dataset similarity learning} that has abundant data
and can be used as an auxiliary meta-task to learn the meta-feature
extractor. The meta-task consists of deciding if two subsets of datasets, where
instances, predictors, and targets have been subsampled,
  so-called multi-fidelity subsets~(\cite{falkner2018bohb}),
belong to the same dataset or not. 
  Each subset is considered an approximation of the entire
  dataset that varies in degree of fidelity depending on the size of the
  subset. In other words, we assume a dataset is similar to a variant of
  itself with fewer instances, predictors, or targets
  (Section~\ref{section: sml}).
  
Finally, we experimentally demonstrate the usefulness of the meta-feature extractor
Dataset2Vec by the correlation of the extracted meta-features with meta-targets of interesting
meta-tasks (\textbf{D4}). Here, we choose \textbf{hyperparameter optimization as the meta-task} (Section~\ref{section: experiments}).

\begin{figure}[h]
\includegraphics[width=1\textwidth]{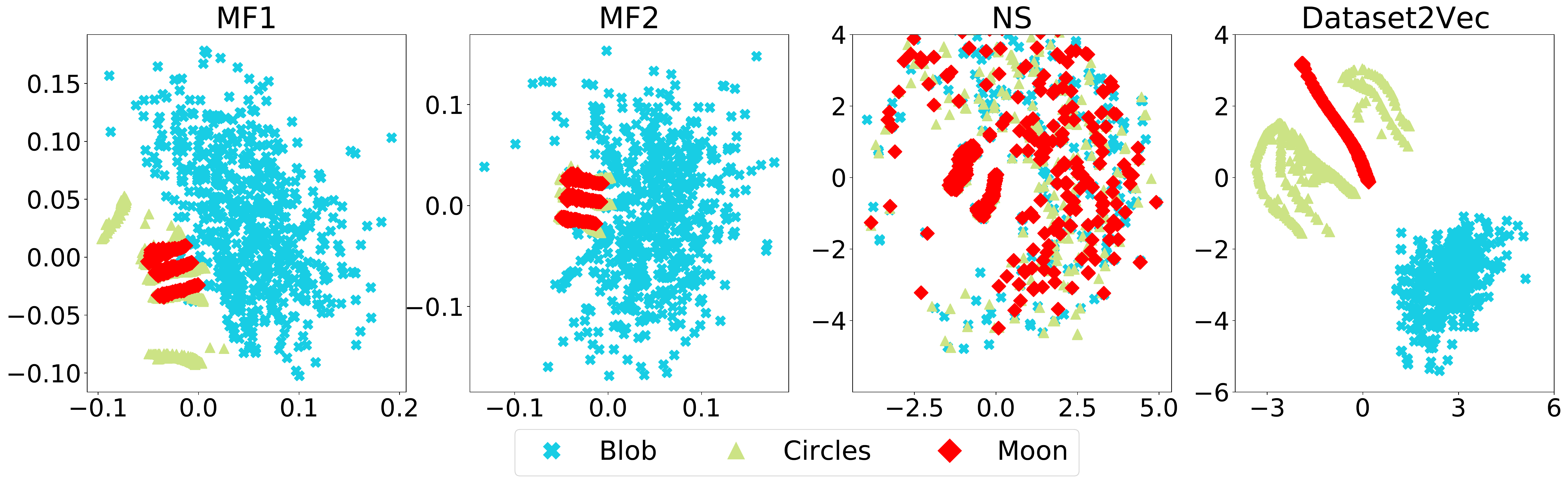}
\caption{Meta-features of 2000 toy datasets extracted by (from left to right)
    engineered dataset meta-features MF1~(\cite{DBLP:conf/pkdd/WistubaSS16}), MF2~(\cite{DBLP:conf/aaai/FeurerSH15}),
    a state-of-the-art model based on variational autoencoders, the Neural Statistician~(\cite{DBLP:conf/iclr/EdwardsS17}), and
    the proposed meta-feature extractor Dataset2Vec.
    The methods compute 22, 46, 64, and 64 meta-features respectively. 
    Depicted is their 2D embedding using multi-dimensional scaling.(Best viewed in color)}
    \label{fig0}
\end{figure}

A way more simple, unsupervised plausibility argument for the usefulness of the extracted
meta-features is depicted in Figure~\ref{fig0} showing a 2D embedding of the meta-features of
2000 synthetic classification toy datasets of three different types (circles/moon/blobs)
computed by
  a) two sets of engineered dataset meta-features: MF1~(\cite{DBLP:conf/pkdd/WistubaSS16}) and MF2~(\cite{DBLP:conf/aaai/FeurerSH15}) (see Table~\ref{char});
  b) a state-of-the-art model based on variational autoencoders, the Neural Statistician~(\cite{DBLP:conf/iclr/EdwardsS17}), and
  c) the proposed meta-feature extractor Dataset2Vec.
For the 2D embedding, multi-dimensional scaling has been applied~(\cite{borg2003modern}) on these meta-features.
As can be clearly seen, the meta-features extracted by Dataset2Vec allow us to separate the three
different dataset types way better than the other two methods (see Section~\ref{section: hpo} for further details).

To sum up, in this paper we make the following key contributions:
\begin{enumerate}
\item We formulate a new problem setting, meta-feature learning for
   datasets with varying schemas.
\item We design and investigate a meta-feature extractor, Dataset2Vec, based on
    a representation of datasets as hierarchical sets of predictor/target pairs.
\item We design a novel meta-task called dataset similarity learning that has abundant
    data and is therefore well-suited as an auxiliary meta-task to train the meta-feature extractor Dataset2Vec.
\item We show experimentally that using the meta-features extracted through
  Dataset2Vec for the hyperparameter optimization meta-task 
  outperforms the use of engineered
  meta-features specifically designed for this meta-task.
\end{enumerate}

\section{Related Work}
In this section, we attempt to summarize some of the topics that relate to our work and highlight where some of the requirements mentioned earlier are (not) met.\\
\textbf{Meta-feature engineering}. Meta-features represent measurable properties of tasks or datasets and play an essential role in meta-learning. Engineered meta-features can be represented as simple statistics~(\cite{DBLP:journals/paa/ReifSGBD14,DBLP:conf/hais/SegreraPM08}) or even as model-specific parameters with which a dataset is trained~(\cite{filchenkov2015datasets}) and are generally applicable to any dataset, schema-agnostic~\textbf{D1}. In addition to that, the nature of these meta-features makes them scalable (\textbf{D3}), and thus can be extracted without extra training. For example, the mean of the predictors can be estimated regardless of the number of targets. However, coupling these meta-features with a meta-task is a tedious process of trial-and-error, and must be repeated for every meta-task to find expressive (\textbf{D2}) meta-features with good correlation (\textbf{D4}) to the meta-target.\\
\textbf{Meta-feature learning}, as a standalone task, i.e. agnostic to a pre-defined meta-task, to the best of our knowledge, is a new concept, with existing solutions bound by a fixed dataset schema. Autoencoder based meta-feature extractors such as the neural statistician (NS)~(\cite{DBLP:conf/iclr/EdwardsS17}) and its variant~(\cite{DBLP:conf/uai/HewittNGJT18}) propose an extension to the conventional variational autoencoder~(\cite{DBLP:journals/corr/KingmaW13}), such that the item to be encoded is the dataset itself. Nevertheless, these techniques require vast amounts~(\cite{DBLP:conf/iclr/EdwardsS17}) of data and are limited to datasets with similar schema, i.e. not schema-agnostic~(\textbf{D2}).\\
\textbf{Embedding and Metric Learning Approaches} aim at learning semantic distance measures that position similar high-dimensional observations within proximity to each other on a manifold, i.e. the meta-feature space. By transforming the data into embeddings, simple models can be trained to achieve significant performance~(\cite{DBLP:conf/nips/SnellSZ17,DBLP:journals/ijon/BerlemontLDG18}). Learning these embeddings involves optimizing a distance metric~(\cite{DBLP:conf/cvpr/SongXJS16}) and making sure that local feature similarities are observed~(\cite{DBLP:journals/tomccap/ZhengZY18}). This leads to more expressive (\textbf{D2}) meta-features that allow for better distinction between observations.\\
\textbf{Meta-Learning} is the process of learning new tasks by carrying over findings from previous tasks based on defined similarities between existing meta-data. Meta-learning has witnessed great success in domain adaptation, learning scalable internal representations of a model by quickly adapting to new tasks~(\cite{DBLP:conf/icml/FinnAL17,DBLP:conf/nips/FinnXL18,DBLP:conf/nips/YoonKDKBA18}). Existing approaches learn generic initial model parameters through sampling tasks from a task-distribution with an associated train/validation dataset. Even within this line of research, we notice that learning meta-features helps achieve state-of-the-art performances~(\cite{DBLP:journals/corr/abs-1807-05960}), but do not generalize beyond dataset schema~(\cite{DBLP:journals/corr/abs-1902-03545,koch2015siamese}). However, potential improvements have been shown with schema-agnostic model initialization~(\cite{brinkmeyer2019chameleon}). Nevertheless, existing meta-learning approaches result in task-dependent meta-features, and hence the meta-features only correlate (\textbf{D4}) with the respective meta-task.\\

We notice that none of the existing approaches that involve meta-features fulfills the complete list of desiderata. As a proposed solution, we present a novel meta-feature extractor, Dataset2Vec, that learns to extract expressive~(\textbf{D2}) meta-features directly from the dataset. Dataset2Vec, in contrast to the existing work, is schema-agnostic (\textbf{D1}) that does not need to be adjusted for datasets with different schema. We optimize Dataset2Vec by a novel dataset similarity learning approach, that learns expressive~(\textbf{D3}) meta-features that maintain inter-dataset and intra-dataset distances depending on the degree of dataset similarities. Finally, we demonstrate the correlation (\textbf{D4}) between the meta-features and unseen meta-tasks, namely hyperparameter optimization, as compared to engineered meta-features. \\
%


\newcommand\mm{\hat y\meta}  
\newcommand\mmmf{\hat Y\meta}  
\newcommand\ymeta{y\meta}  
\newcommand\ymetahat{\hat y\meta}  

\section{Problem Setting: Meta-feature Learning}
\label{section: mfl}
A (supervised) learning task is usually understood as
a problem to find a function (model) that maps given predictor
values to likely target values based on past observations
of predictors and associated target values (dataset).
Many learning tasks depend on further inputs besides
the dataset, e.g., hyperparameters like the depth and width
of a neural network model, a regularization weight, 
a specific way to initialize the parameters of a model, etc.
These additional inputs of a learning task often are
found heuristically, e.g., hyperparameters can be found
by systematic grid or by random search~(\cite{bergstra2012random}), model parameters
can be initialized by random normal samples, etc.
From a machine learning perspective, finding these additional
inputs of a learning task can itself be described as a learning task:
its inputs are a whole dataset, its output is a hyperparameter
vector or an initial model parameter vector. To differentiate
these two learning tasks we call the first task, to learn a model
from a dataset, the \textbf{primary learning task}, and the second,
to find good hyperparameters or a good model initialization,
the \textbf{meta-learning task}. Meta-learning tasks are very special
learning tasks as their inputs are not
  simple vectors like in traditional classification and regression tasks, nor
  sequences or images like in time-series or image classification,
but
  themselves whole datasets.

To leverage standard vector-based machine learning models
for such meta-learning tasks, their inputs, a whole dataset, must be
described by a vector. Traditionally, this vector is engineered by experts and contains simple statistics
such as the number of instances, the number of predictors,
the number of targets, the mean and variance of the mean of the
predictors, etc. These vectors that describe whole datasets and that are
the inputs of the meta-task are called \textbf{meta-features}. The meta-features
together with the meta-targets, i.e. good hyperparameter values
or good model parameter initializations for a dataset, form the
\textbf{meta-dataset}.

More formally, let ${\cal D}$ be the space of all possible
datasets,
\begin{align*}
  {\cal D} :=  \{ D \in \R^{N\times (M+T)} \mid N,M,T\in\N\}
\end{align*}
i.e. a data matrix containing a row for each instance and a column for each predictor
and target together with the number $M$ of predictors (just to mark which columns
are predictors, which targets).
For simplicity of notation, for a dataset $D\in{\cal D}$ we will denote
by $N^{(D)}, M^{(D)}$ and $T^{(D)}$ its number of instances, predictors and targets
and by $X^{(D)}, Y^{(D)}$ its predictor and target matrices (see Table~\ref{notations}).
Now a meta-task is a learning task that aims to find a \textbf{meta-model}
$\mm: {\cal D} \rightarrow \R^{T\meta}$, e.g., for hyperparameter learning
of three hyperparameters depth and width of a neural network and
regularization weight, to find good values for each given dataset
(hence here $T\meta=3$), or for model parameter initialization
for a neural network with 1 million parameters, 
to find good such initial values (here $T\meta=1,000,000$).

Most meta-models $\mm$ are the composition of two functions:
\begin{enumerate}
\item[i)] the \textbf{meta-feature extractor} 
    $\hat\phi : {\cal D}\rightarrow \mathbb{R}^K$,
  that extracts from a dataset a fixed number $K$ of meta-features,
and
\item[ii)] a \textbf{meta-feature based meta-model}
    $\mmmf: \mathbb{R}^K \rightarrow \mathbb{R}^{T\meta}$
  that predicts the meta-targets based on the meta-features and
  can be a standard vector-based regression model chosen for the
  meta-task at hand, e.g., a neural network.
\end{enumerate}
Their composition yields the meta-model $\mm$:
\begin{equation}
\label{end-to-end}
  \mm : {\cal D}\overset{\hat\phi}{\longrightarrow}
            \mathbb{R}^K \overset{\mmmf}{\longrightarrow} \mathbb{R}^{T\meta}
\end{equation}

Let $a\meta$ denote the learning algorithm for the meta-feature based meta-model,
i.e. stochastic gradient descent to learn a neural network that predicts
good hyperparameter values based on dataset meta-features.

The \textbf{Meta-feature learning problem} then is as follows:
given
  i) a meta-dataset $(\cal D\meta$,$\Y\meta)$ of pairs of (primary) datasets $D$
    and their meta-targets $\ymeta$,
 ii) a meta-loss $\ell\meta: \Y\meta\times\Y\meta\rightarrow\R$ where
      $\ell\meta(\ymeta,\ymetahat)$ measures how bad the predicted meta-target $\ymetahat$
      is for the true meta-target $\ymeta$, and
 iii) a learning algorithm $a\meta$ for a meta-feature based meta-model
    (based on $K\in\N$ meta-features),
find a meta-feature extractor
$\hat\phi: {\cal D}\rightarrow \R^K$
s.t. the expected meta-loss
   of the meta-model learned by $a\meta$ from the meta-features extracted by $\hat\phi$ 
   over new meta-instances is minimal:
\begin{equation*}
  \min_{\hat \phi}\mathbb{E}_{D,\ymeta}(\ell\meta(\ymeta, \mm(D)))
\end{equation*}
such that:
\begin{equation*}
\begin{split}
      \mm&:= \hat\phi \circ \mmmf\\
      \mmmf&:= a\meta(X\meta, Y\meta)\\
       X\meta&:= ( \hat\phi(D))_{D\in D\meta}
\end{split}
\end{equation*}

Different from standard regression problems where the loss
is a simple distance between true and predicted targets such as
the squared error, the meta-loss is more complex as
its computation involves a primary model being learned and evaluated for the primary
learning task. For hyperparameter optimization, if the best
depth and width of a neural network for a specific task is $10$ and $50$,
it is not very meaningful to measure the squared error distance to a predicted
depth and width of say $5$ and $20$. Instead one is interested in the
difference of primary test losses of primary models learned with these
hyperparameters. So, more formally again, let $\ell$ be the primary loss,
say squared error for a regression problem, and $a$ be a learning
algorithm for the primary model, then
\begin{equation*}
\begin{split}
   \ell\meta( \ymeta, \ymetahat) &:=  \mathbb{E}_{x,y} \ell(y, \hat y^*(x)) - \mathbb{E}_{x,y} \ell(y, \hat y(x))\\
    \hat y^*&:= a(X^{(D)},Y^{(D)}, \ymeta)\\ \hat y&:= a(X^{(D)},Y^{(D)}, \ymetahat)
\end{split}    
\end{equation*}

\commentout{
Meta-features represent dataset descriptions used to accelerate
learning across tasks or datasets and are either engineered or learned
as part of the meta-task. Let $\cal M$ be a meta-feature-based
meta-model that solves a specified meta-task, where
  $\cal M : {\cal X}\meta\rightarrow\Y\meta$,
such that
  ${\cal X}\meta\in\mathbb{R}^K$ represents the meta-feature input and
  $\Y\meta\in\mathbb{R}$ represents the output.
Ideally, we can integrate the meta-feature extractor,
  $\phi : D\rightarrow \mathbb{R}^K$,
into the model and train it simultaneously, i.e. end-to-end, instead of
relying on engineered meta-features. Assuming, without loss of
generality, that the meta-feature extractor and the meta-model
are parameterized by neural networks, the end-to-end algorithm
is formulated in Equation~\ref{end-to-end}:
\begin{equation}
\label{end-to-end}
  \cal M : \text{$D$}\rightarrow \mathbb{R}^\text{k} \rightarrow \Y\meta
\end{equation}
But most meta-learning datasets are small, rarely containing more than
a couple of thousands or 10,000s of meta-instances, as each such
meta-instance itself requires an independent target learning
process. Thus training a meta-feature extractor end-to-end directly,
is prone to overfitting.

We propose to employ an auxiliary meta-task, with abundant data, to
extract meaningful meta-features. More specifically, we define our
meta-task as a dataset similarity learning problem, where the
objective is to enforce the proximity of meta-features between similar
datasets and increase the distance between the meta-features of
dissimilar datasets. However, the lack of explicit feedback regarding
the similarity of tabular datasets remains an open challenge in
meta-feature learning. To circumvent this problem, we rely on
implicit feedback by defining similar datasets as variants of the same
dataset.
}

\newcommand\ofD{^{\left(D\right)}}
\section{The Meta-feature Extractor Dataset2Vec}
\label{section: d2v}

To define a learnable meta-feature extractor
  $\hat\phi : \cal D\rightarrow \mathbb{R}^K$,
we will employ the Kolmogorov-Arnold representation theorem~(\cite{kuurkova1991kolmogorov})
and use the DeepSet architecture~(\cite{DBLP:conf/nips/ZaheerKRPSS17}).


\begin{table}[t]
	\centering
	\begin{tabular}{D{+}{\,\in\,}{-1} l}
		\toprule
		\multicolumn{1}{c}{\textbf{Notation}} & \textbf{Description} \\ \midrule
		D\;\;\; + {\cal D} & A dataset $D$ in the datasets' space $\cal D$ \\ 
		N\ofD + \N & Number of instances in dataset $D$ \\
		M\ofD + \N & Number of predictors in $D$ \\
		T\ofD + \N & Number of classes/targets in dataset $D$ \\
		X\ofD + \R^{N\ofD\times M\ofD} & Predictors of dataset $D$ \\ 
		Y\ofD+\R^{N\ofD\times T\ofD} & Targets of dataset $D$  \\
		\hat\phi\left(D\right)+\R^K & The $K$-dimensional meta-features of $D$ \\ \bottomrule 
	\end{tabular}
\caption{Notations}
\label{notations}
\end{table}

\subsection{Preliminaries}
The Kolmogorov-Arnold representation theorem~(\cite{kuurkova1991kolmogorov}) states that any multivariate function $\phi$ of $M$-variables $X_1,\dots,X_M$ can be represented as an aggregation of univariate functions~(\cite{kuurkova1991kolmogorov}), as follows:

\begin{equation}
\phi\left(X_1, \dots, X_M \right) \approx \sum_{k=0}^{2M} h_k\left( \sum_{\ell=1}^{M} g_{\ell,k}\left(X_l\right) \right)
\end{equation}

The ability to express a multivariate function $\phi$ as an aggregation $h$ of single variable functions $g$ is a powerful representation in the case where the function $\phi$ needs to be invariant to the permutation order of the inputs $X$. In other words, $\phi\left(X_1,\dots,X_M\right) = \phi\left(X_{\pi(1)},\dots,X_{\pi(M)}\right)$ for any index permutation $\pi(k)$, where $k \in \{1,\dots, M\}$. To illustrate the point further with $M=2$, we would like a function $\phi(X_1, X_2)=\phi\left(X_2,X_1\right)$ to achieve the same output if the order of the inputs $X_1, X_2$ is swapped. In a multi-layer perceptron, the output changes depending on the order of the input, as long as the values of the input variables are different. The same behavior is observed with recurrent neural networks and convolutional neural networks.

However, permutation invariant representations are crucial if functions are defined on sets, for instance if we would like to train a neural network to output the sum of a set of digit images. Set-wise neural networks have recently gained prominence as the Deep-Set formulation~(\cite{DBLP:conf/nips/ZaheerKRPSS17}), where the $2M+1$ functions $h$ of the original Kolmogorov-Arnold representation is simplified with a single large capacity neural network $h \in \R^K \rightarrow \R$, and the $M$-many univariate functions $g$ are modeled with a shared cross-variate neural network function $g \in \R \rightarrow\R^K$:

\begin{equation}
\phi\left(X_1, \dots, X_M \right) \approx h\left(\sum_{k=1}^{M} g\left(X_k\right) \right)
\end{equation}

Permutation-invariant functional representation is highly relevant for deriving meta-features from tabular datasets, especially since we do not want the order in which the predictors of an instance are presented to affect the extracted meta-features.
\subsection{Hierarchical Set Modeling of Datasets}
\label{base-mfe}

In this paper, we design a novel meta-feature extractor called Dataset2Vec for tabular datasets as a hierarchical set model. Tabular datasets are two-dimensional matrices of $(\#\text{rows} \times \#\text{columns})$, where the columns represent the predictors and the target variables and the rows consist of instances. As can be trivially understood, the order/permutation of the columns is not relevant to the semantics of a tabular dataset, while the rows are also permutation invariant due to the identical and independent distribution principle. In that perspective, a tabular dataset is a set of columns (predictors and target variables), where each column is a set of row values (instances):

\begin{itemize}
	\renewcommand{\labelitemi}{$\square$} 
	\renewcommand{\labelitemii}{$\square$}
	\renewcommand{\labelitemiii}{$\square$}
	\renewcommand{\labelitemiv}{$\square$}
	\item A dataset is a set of $M\ofD+T\ofD$ predictor and target variables:
	\begin{itemize}
		\item $D = \left\{X\ofD_1, \dots, X\ofD_{M\ofD},Y\ofD_1, \dots, Y\ofD_{T\ofD}\right\}$
			\begin{itemize}
			\item Where each variable is a set of $N\ofD$ instances:
				\begin{itemize}
				\item $X\ofD_m = \left\{X\ofD_{1,m}, \dots, X\ofD_{N\ofD,m}\right\}$, $m=1,\dots,M\ofD$
				\item $Y\ofD_t = \left\{Y\ofD_{1,t}, \dots, Y\ofD_{N\ofD,t}\right\}$, $t=1,\dots,T\ofD$
			\end{itemize}
		\end{itemize}
	\end{itemize}
\end{itemize}

\begin{figure}[t]
  \centering
  \subfloat{\includegraphics[width=1\columnwidth]{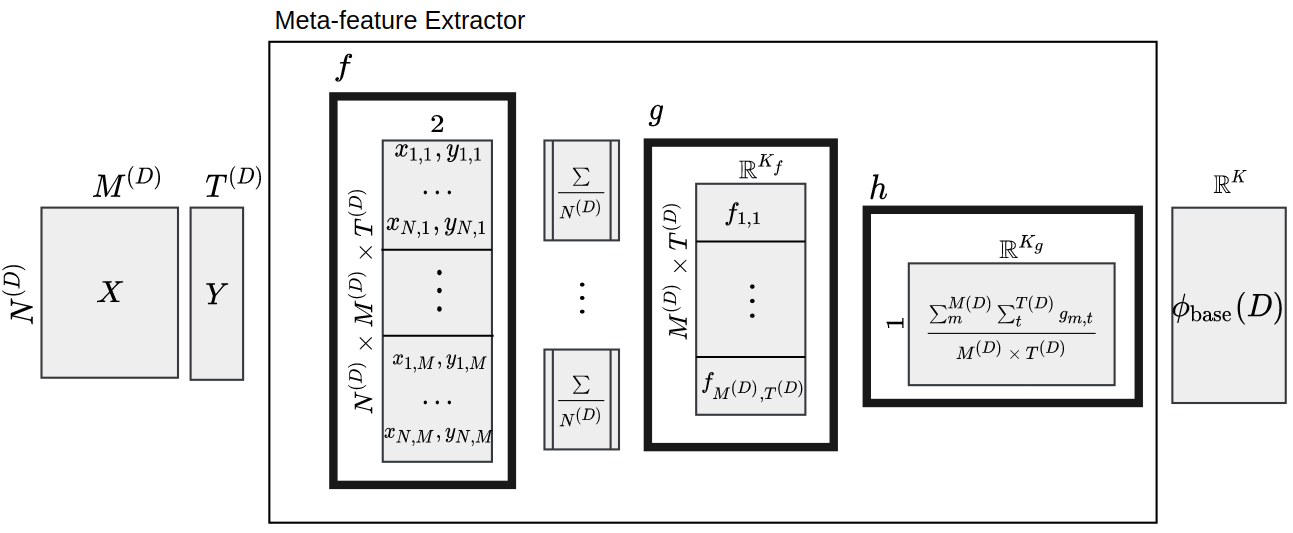}}
  \caption{Overview of the Dataset2Vec as described in Section~\ref{base-mfe}}  
  \label{fig: mfe ex}
\end{figure}

In other words, a dataset is a \textbf{set of sets}. Based on this novel conceptualization we propose to model a dataset as a hierarchical set of two layers. More formally, let us restate that $\left(X\ofD,Y\ofD\right) = D\in{\cal D}$ is a dataset, where
$X\ofD\in\R^{N\ofD\times M\ofD}$ with $M\ofD$ represents the number of predictors and $N\ofD$ the number of instances, and $Y\ofD\in\R^{N\ofD\times T\ofD}$ with $T\ofD$ as the number of targets.
We model our meta-feature extractor, Dataset2Vec, without loss of generality, as a feed-forward neural network, which accommodates all schemas of datasets. Formally, Dataset2Vec is defined in Equation~\ref{eq1}:
\begin{equation} \label{eq1}
\hat\phi(D)  := h\left(\frac{1}{M\ofD T\ofD}\sum_{m=1}^{M\ofD}\sum_{t=1}^{T\ofD}g \left(\frac{1}{N\ofD}\sum_{n=1}^{N\ofD}f\big(X\ofD_{n,m},Y\ofD_{n,t}\big)\right) \right)
\end{equation}
with
  $f: \R^2\rightarrow\R^{K_f}$,
  $g: \R^{K_f}\rightarrow\R^{K_g}$ and
  $h: \R^{K_g}\rightarrow\R^{K}$
represented by neural networks with $K_f$, $K_g$ and $K$ output units, respectively. 
Notice that the best way to design a scalable meta-feature extractor is by reducing the input to a single predictor-target pair $\left(X\ofD_{n,m}, Y\ofD_{n,t}\right)$. This is especially important to capture the underlying correlation between each predictor/target variable.
Each function is designed to model a different aspect of the dataset, instances, predictors, and targets. Function $f$
  captures the interdependency between an instance feature $X\ofD_{n,m}$ and the corresponding instance target $Y\ofD_{n,t}$ followed a pooling layer across all instances $n\in N\ofD$. Function $g$ extends the model across all targets $t\in T\ofD$ and predictors $m\in M\ofD$. Finally, function $h$ applies a transformation to the average of latent representation collapsed over predictors and targets, resulting in the meta-features. Figure~\ref{fig: mfe ex} depicts the architecture of Dataset2Vec.

\subsection{Network Architecture}
Our Dataset2Vec architecture is divided into three modules, $\hat{\phi}:= f\circ g\circ h$, each implemented as neural network. Let \textbf{Dense(n)} define one fully connected layer with $n$ neurons, and \textbf{ResidualBlock(n,m)} be $m\times$ Dense(n) with residual connections~(\cite{DBLP:conf/bmvc/ZagoruykoK16}). We present two architectures in Table~\ref{table: network architecture}, one for the toy meta-dataset, Section~\ref{subsection: toy}, and a deeper one for the tabular meta-dataset, Section~\ref{subsection: uci}.
All layers have Rectified Linear Unit activations (ReLUs). Our reference implementation uses Tensorflow~(\cite{DBLP:conf/osdi/AbadiBCCDDDGIIK16}).
\begin{table}[h]
\caption{Network Architectures.}
\centering
\scalebox{1.0}{
\begin{tabular}{ll}
\toprule
Functions &    Toy Meta-dataset Architecture\\
\midrule
$f$      &  Dense(64);ResidualBlock(3,64);Dense(64)\\
$g$      &  2$\times$Dense(64)\\
$h$      &  Dense(64);ResidualBlock(3,64);Dense(64)\\
\bottomrule
Functions &    Tabular Meta-dataset Architecture\\
\midrule
$f$      &  7$\times$[Dense(32);ResidualBlock(3,32);Dense(32)]\\
$g$      &  Dense(32);Dense(16);Dense(8)\\
$h$      &  3$\times$[Dense(16);ResidualBlock(3,16);Dense(16)]\\
\bottomrule
\end{tabular}}
\label{table: network architecture}
\end{table}


\section{The Auxiliary Meta-Task: Dataset Similarity Learning}
\label{section: sml}

Ideally, we can train the composition $\hat M\circ\hat \phi$ of meta-feature
extractor $\hat\phi$ and meta-feature based meta-model $\hat M$
end-to-end.
But most meta-learning datasets are small, rarely containing more than
a couple of thousands or 10,000s of meta-instances, as each such
meta-instance itself requires an independent primary learning
process. Thus training a meta-feature extractor end-to-end directly,
is prone to overfitting.
Therefore, we propose to employ additionally an auxiliary meta-task
with abundant data to extract meaningful meta-features.

\subsection{The Auxiliary Problem}
\textbf{Dataset similarity learning} is the following
novel, yet simple meta-task:
given a pair of datasets and an assigned dataset similarity indicator $(x,x',i)\in{\cal D\meta}\times{\cal D\meta}\times\{0,1\}$
from a joint distribution $p$ over datasets,
learn a dataset similarity learning model $\hat i: {\cal D\meta}\times{\cal D\meta}\rightarrow\{0,1\}$
with minimal expected misclassification error 
\begin{equation}
   E_{(x,x',i)\sim p}( I(i \neq \hat i(x,x')))
\end{equation}
where $I(\text{true}):=1$ and $I(\text{false}):=0$.

\begin{figure}[h]
	\begin{minipage}[c]{0.5\textwidth}
		\includegraphics[width=0.7\textwidth, trim={0cm, 0cm, 3cm, 0cm}]{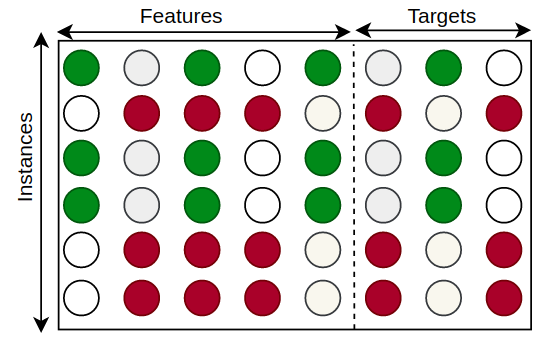}
	\end{minipage}\hfill
	\begin{minipage}[c]{0.5\textwidth}
		\caption{Illustrating Algorithm~\ref{alg:alg1}:
			Two subsets $D_s$ in \textcolor{green}{green} and $D'_s$ in \textcolor{red}{red} are drawn randomly from the same dataset and annotated as $i(D_s, D'_s) = 1$.}
		\label{fig:batch sampler dataset id problem}
	\end{minipage}
\end{figure}

As mentioned previously, learning meta-features from datasets $D\in{\cal D\meta}$ directly
is impractical and does not scale well to larger datasets, especially due to the lack of an explicit dataset similarity label. To overcome this limitation, we create implicitly similar datasets in the form of multi-fidelity subsets of the datasets, batches, and assign them a dataset similarity label $i:=1$,  whereas variants of different datasets are assigned a dataset similarity label $i:=0$.
Hence, we define the multi-fidelity subsets for any specific dataset $D$ as the submatrices pair $(X',Y')$, Equation~\ref{batch}:
\begin{equation}
\label{batch}
    X':= (X\ofD_{n,m})_{n\in N',m\in M'}, \quad Y':= (Y\ofD_{n,t})_{n\in N',t\in T'}
\end{equation}
with $N'\subseteq\{1,\ldots,N\ofD\}$, $M'\subseteq\{1,\ldots,M\ofD\}$, and $T'\subseteq\{1,\ldots,T\ofD\}$
representing the subset of indices of instances, features, and targets, respectively, sampled 
from the whole dataset.

The batch sampler, Algorithm~\ref{alg:alg1}, returns a batch with 
random index subsets $N'$,$M'$ and $T'$ drawn uniformly from
$\{2^q | q \in [4,8]\}, [1,M]$ and $[1,T]$,
without replacement. Figure~\ref{fig:batch sampler dataset id problem} is a pictorial
representation of two randomly sampled batches from a tabular dataset.

\begin{algorithm}\caption{$\textbf{sample-batch}(D)$}\label{alg:alg1}

\begin{algorithmic}[1]
\STATE{\algorithmicrequire Dataset $D$}
\STATE{$N' \sim \{2^q | q \in [4,8]\}$}
\STATE{$M' \sim [1,M\ofD]$}
\STATE{$T' \sim [1,T\ofD]$}
\STATE{$X':= (X\ofD_{n,m})_{n\in N',m\in M'}$}
\STATE{$Y':= (Y\ofD_{n,t})_{n\in N',t\in T'}$}
\RETURN $(X',Y')$
\end{algorithmic}
\end{algorithm}

Meta-features of a dataset $D$ can then be computed either directly over the dataset as a whole, $\hat\phi(D)$, or estimated as the 
average of several random batches, Equation~\ref{mfe}:
\begin{equation}
\label{mfe}
  \hat\phi(D):= \frac{1}{B} \sum_{b=1}^{B} \hat\phi(\text{sample-batch}(D))
\end{equation}
where $B$ represents the number of random batches.

Let $p$ be any distribution on pairs of dataset batches and $i$ a
binary value indicating if both subsets are similar, then given a
distribution of data sets $p_{\cal D\meta}$, we sample multi-fidelity
subset pairs for the dataset similarity learning problem using
Algorithm~\ref{alg:alg2}.

\begin{algorithm}\caption{$\textbf{sample-batch-pairs}(D)$}\label{alg:alg2}
\begin{algorithmic}[1]
\STATE{\algorithmicrequire Distribution over datasets $p_{\cal D\meta}$}
\STATE{$D \sim p_{\cal D\meta}$}
\IF{$\text{unif}([0,1]) < 0.5$}
\STATE{$D' \sim p_{{\cal D\meta}\setminus D}$}
\STATE{$i:= 0$}
\ELSE
\STATE{$D':= D$}
\STATE{$i:= 1$}
\ENDIF
\STATE{$D_s:= \text{sample-batch}(D)$}
\STATE{$D'_s:= \text{sample-batch}(D')$}
\RETURN $(D_s, D'_s, i)$
\end{algorithmic}
\end{algorithm}

\subsection{The Auxiliary Meta Model and Training}
To force all information relevant for the dataset similarity learning task to be pooled
in the meta-feature space, we use a probabilistic dataset similarity learning model, Equation~\ref{non-parametric task}:
\begin{equation}
\label{non-parametric task}
  \hat i(x, x'):= e^{-\gamma||\hat\phi(x)-\hat\phi(x')||}
\end{equation}
with $\gamma$ as a tuneable hyperparameter.
We define pairs of similar batches as $P=\{(x,x',i) \sim p{}|{}i = 1\}$ and pairs of dissimilar batches $W=\{(x,x',i) \sim p{}|{}i = 0\}$, and formulate the optimization objective as:
\begin{equation}
\text{arg}\min_{\hat\phi} -\frac{1}{|P|}\sum_{(x,x',i)\in P} \log(\hat i(x,x')) - \frac{1}{|W|}\sum_{(x,x',i)\in W}\log(1-\hat i(x,x'))
\end{equation}
Similar to any meta-learning task, we split the meta-dataset, into ${\cal D\meta}_{train}$, ${\cal D\meta}_{valid}$ and ${\cal D\meta}_{test}$
which include non-overlapping datasets for training, validation and test respectively.
 While training the dataset similarity learning task, all the latent information are captured in the final layer of Dataset2Vec, our meta-feature extractor, 
 resulting in task-agnostic meta-features. The pairwise loss between batches allows to preserves the intra-dataset
similarity, i.e. close proximity of meta-features of similar batches, as well as inter-dataset similarity, i.e. distant meta-features of dissimilar batches.
\\
Dataset2Vec is trained on a large number of batch samples from datasets in a meta-dataset,
thus currently does not use any information of the subsequent meta-problem to solve and hence
is generic, in this sense unsupervised meta-feature extractor. Any other meta-task
could be easily integrated into Dataset2Vec by just learning them jointly in a multi-task
setting, especially for dataset reconstruction similar to the dataset reconstruction of
the NS~(\cite{DBLP:conf/iclr/EdwardsS17}) could be interesting if one could
figure out how to do this across different schemata. We leave this aspect for future work.

\section{Experiments}
\label{section: experiments}
We claim that for a meta-feature extractor to produce useful meta-features it must meet the following requirements: schema-agnostic (\textbf{D1}), expressive (\textbf{D2}), scalable (\textbf{D3}), and correlates to meta-tasks (\textbf{D4}). We train and evaluate Dataset2Vec in support of these claims by designing the following two experiments. Accordingly, we highlight where the criterion is met throughout the section. Implementation can be found here\footnote{https://github.com/hadijomaa/dataset2vec.git}.
\subsection{Dataset Similarity learning for datasets of similar schema}
\label{subsection: toy}
Dataset similarity learning is designed as an auxiliary meta-task that allows Dataset2Vec to capture all the required information in the meta-feature space to distinguish between datasets. We learn to extract expressive meta-features by minimizing the distance between meta-features of subsets from similar datasets and maximizing the distance between the meta-features of subsets from different datasets. We stratify the sampling during training to avoid class imbalance. The reported results represent the average of a 5-fold cross-validation experiment, i.e. the reported meta-features are extracted from the datasets in the meta-test set to illustrate the scalability (\textbf{D3}) of Dataset2Vec to unseen datasets.

\subsubsection*{Baselines}
We compare with the neural statistician algorithm, NS~(\cite{DBLP:conf/iclr/EdwardsS17}), a meta-feature extractor that learns meta-features as context information by encoding complete datasets with an extended variational autoencoder. We focus on this technique particularly since the algorithm is trained in an unsupervised manner, with no meta-task coupled to the extractor. However, since it is bound by a fixed dataset schema, we generate a 2D labeled synthetic (toy) dataset, to fairly train the spatial model presented by the authors\footnote{https://github.com/conormdurkan/neural-statistician.git}. We use the same hyperparameters used by the authors. We also compare with two sets of well-established engineered meta-features: MF1~(\cite{DBLP:conf/pkdd/WistubaSS16}) and MF2~(\cite{DBLP:conf/aaai/FeurerSH15}). A brief overview of the meta-features provided by these sets is presented in Table~\ref{char}. For detailed information about meta-features, we refer the readers to~(\cite{DBLP:journals/corr/abs-1808-10406}).
\subsubsection*{Evaluation Metric}
We evaluate the similarity between embeddings through pairwise classification accuracy with a cut-off threshold of $\frac{1}{2}$. We set the hyperparameter $\gamma$ in Equation~\ref{non-parametric task} to $1$,$0.1$, and $0.1$ for MF1, MF2, and NS, respectively, after tuning it on a separate validation set, and keep $\gamma = 1$ for Dataset2Vec.
We evaluate the pairwise classification accuracy over 16,000 pairs of batches containing an equal number of positive and negative pairs.
The results are a summary of 5-fold cross validation, during which the test datasets are not observed in the training process.
\begin{table}[h]
\caption{A sample overview of the engineered meta-features.}
\centering
\begin{center}
\scalebox{0.9}{
\begin{tabular}{lll}
\toprule
Method Name & Count & Description\\
\midrule
MF1   & 22 & Kurtosis, Skewness, Class probability, etc.~(\cite{DBLP:conf/pkdd/WistubaSS16})\\
MF2   & 46 & Landmarking, PCA, Missing values, etc.~(\cite{DBLP:conf/aaai/FeurerSH15})\\
D2V   & 64 & -\\
\bottomrule
\end{tabular}}
\end{center}
\label{char}
\end{table}
\subsubsection*{Toy Meta Dataset}
\begin{figure}[h]
    \centering
	\subfloat{\includegraphics[width=0.7\columnwidth]{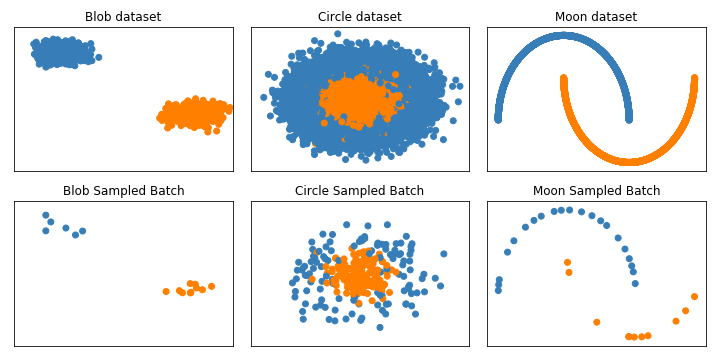}}
	\caption{An example of the 2D toy meta-datasets generated for dataset similarity learning.}
	\label{fig: synthetic data}
\end{figure}

We generate a collection of 10,000 2D datasets each containing a varying number of samples. The datasets are created using the sklearn library~(\cite{scikit-learn}) and belong to either circles or moons with 2 classes (default), or blobs with varying number of classes drawn uniformly at random within the bounds $(2,8)$. We also perturb the data by applying random noise. The toy meta-dataset is obtained by Algorithm~\ref{alg:alg3}. An example of the resulting datasets is depicted, for clarity, in Figure~\ref{fig: synthetic data}.

\begin{algorithm}[h]
\begin{algorithmic}[1]

	\STATE{\algorithmicrequire Number of Features $M$}
	\STATE{	random state $s \sim [0,100]$}
	\STATE{ number of instances $N \sim \{2^q|q\in [11,14]\}$}
	\STATE{ type of dataset $ds \sim \{\text{circles,blobs,moons}\}$}
	\STATE{$X,Y := \text{make}\_ds(N,s,M)$}
   	\IF{$\text{unif}[0,1] < 0.5$}
   	\STATE{$X := apply\_noise(X)$}
   	\ENDIF
	\RETURN $X,Y$
\end{algorithmic}
	\caption{$\textbf{generate-set}(M)$}
	\label{alg:alg3}
\end{algorithm}

We randomly sample a fixed-size subset of 200 samples from every dataset for both approaches, adhering to the same conditions in NS to ensure a fair comparison, and train both networks until convergence. We also extract the engineered meta-features from these subsets. The pairwise classification accuracy is summarized in Table~\ref{res}. We conducted a T-test to validate the distinction between the performance of Dataset2Vec and MF1, the second-best performing approach. The test is a standard procedure to compare the statistical difference of methods which are run once over multiple datasets~(\cite{demvsar2006statistical}). Dataset2Vec has a statistical significance p-value of $3.25\times10^{-11}$, hence significantly better than MF1, following a 2-tailed hypothesis with a significance level of 5\% (standard test setting). Dataset2Vec, compared to NS, has $45\times$ fewer parameters in addition to learning more expressive meta-features.
\begin{table}[h]
\caption{Pairwise Classification Accuracy}
\centering
\begin{center}
\scalebox{0.9}{
\begin{tabular}{lrr}
\toprule
Method & Nb Parameters & Accuracy (\%)\\
\midrule
MF1~(\cite{DBLP:conf/pkdd/WistubaSS16})&-& 85.20 $\pm$  0.35\\
MF2~(\cite{DBLP:conf/aaai/FeurerSH15})&-& 67.82 $\pm$  0.41\\
NS~(\cite{DBLP:conf/iclr/EdwardsS17})&2271402& 57.70 $\pm$  0.93\\
\hline
Dataset2Vec & 50112 & 96.19 $\pm$  0.28 \\
\bottomrule
\end{tabular}}
\end{center}
\label{res}
\end{table}

\begin{figure}[h]
    \centering
	\subfloat{\includegraphics[width=1\columnwidth]{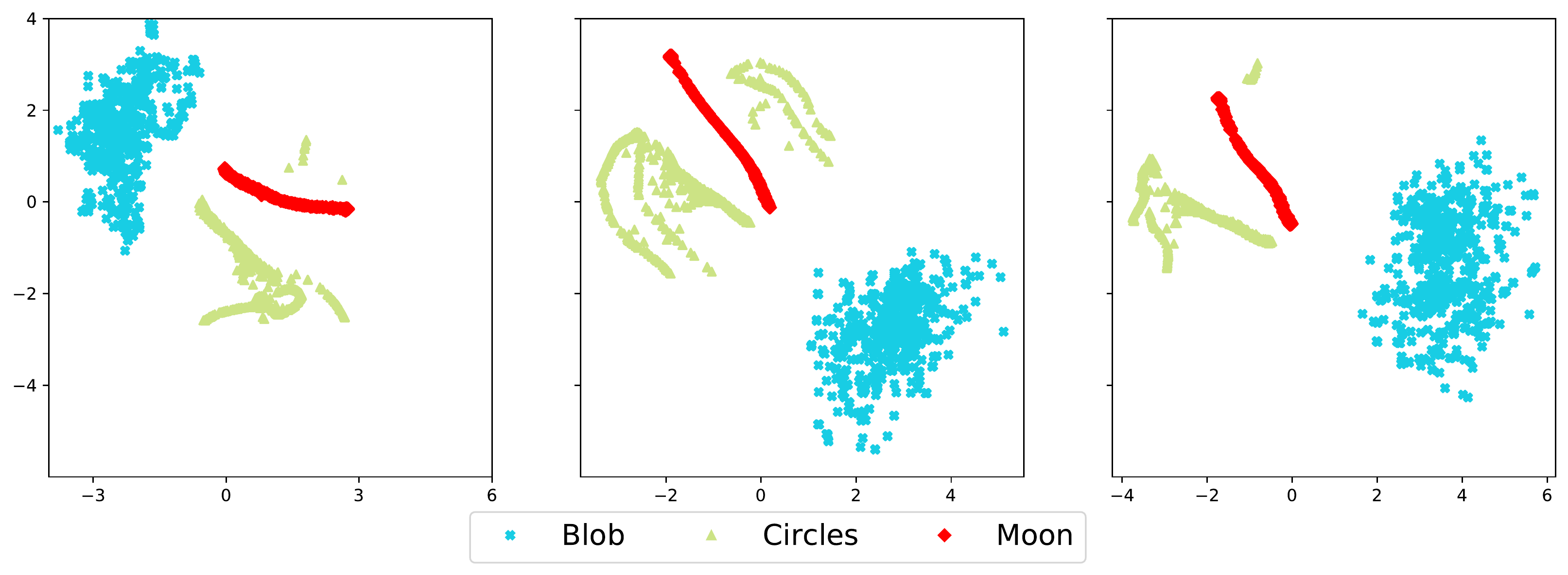}}
	\caption{
			The figure illustrates the 2D projections based on multi-dimensional scaling~(\cite{borg2003modern}) of our learned meta-features for three distinct folds. Each point in the figure represents a single dataset that is either a blob, a moon, or a circle, generated synthetically with different parameters selected from the test set of each fold, i.e. never seen by our model during training. The depiction highlights that Dataset2Vec is capable of generating meta-features from unseen datasets while preserving inter-and intra-dataset similarity. This is demonstrated by the co-location of the meta-features of similar datasets, circles near circles, etc. in this 2D space. (Best viewed in color)}
	\label{fig: toy dataset visual summary}
\end{figure}

The expressivity (\textbf{D2}) of Dataset2Vec is realized in Figure~\ref{fig: toy dataset visual summary} which depicts a 2D projection of the learned meta-features, as we observe collections of similar datasets with co-located meta-features in the space, and meta-features of different datasets are distant.

Intuitively, it is also easy to understand how the datasets of circles and moons, might be more similar compared to blobs, as seen by the projections in Figure~\ref{fig: toy dataset visual summary}. This might be largely attributed to the large distance between instances of dissimilar classes in the blob datasets, whereas instances of dissimilar classes in the moon and the circle datasets are closer, Figure~\ref{fig: synthetic data}.
\subsection{Dataset Similarity learning for datasets of different schema}
\label{subsection: uci}
\label{sec: bi for different schema}
For a given machine learning problem, say binary classification, it is unimaginable that one can find a large enough collection
of similar and dissimilar datasets with the same schema. The schema presents an obstacle that hinders the potential 
of learning useful meta-tasks. Dataset2Vec is schema-agnostic (\textbf{D1}) by design.
\subsubsection*{UCI Meta Dataset}
The UCI repository~(\cite{Dua:2019}) contains a vast collection of datasets. We used 120 preprocessed classification datasets\footnote{http://www.bioinf.jku.at/people/klambauer/data$\_$py.zip} with a varying schema to train the meta-feature extractor by randomly sampling pairs of subsets, Algorithm~\ref{alg:alg2}, along with the number of instances, predictors, and targets. Other sources of tabular datasets are indeed available~(\cite{vanschoren2014openml}),
nevertheless, they suffer from quality issues (missing values, require pre-processing, etc.), which is why 
we focus on pre-processed and normalized UCI classification datasets.\\ We achieve a pairwise classification accuracy of 88.20\% $\pm$  1.67, where the model has 45424 parameters.
In Table~\ref{table: uci dataset anecdotdal summary}, we show five randomly selected groups of datasets that have been collected by a 5-Nearest Neighbor method based on the Dataset2Vec meta-features. We rely on the semantic similarity, i.e. similarity of the names, of the UCI datasets to showcase neighboring datasets
in the meta-feature space due to the \textit{lack of an explicit dataset similarity annotation measure for tabular datasets}.
For this meta-dataset, NS could not be applied due to the varying dataset schema.

\begin{small}
\begin{table}[htp]
\caption{Groups of dataset based on the 5-NN of their meta-features}
\centering
\begin{center}
\scalebox{1}{
\begin{tabular}{lllll}
\toprule
\textbf{Group 1} & \textbf{Group 2} & \textbf{Group 3} & \textbf{Group 4} & \textbf{Group 5} \\
\midrule
  monks-1 &     echocardiogram &            credit-approval &  post-operative &      pb-SPAN  \\
  monks-2 &  heart-switzerland &          australian-credit &   lung-cancer      &    pb-T-OR-D \\
  monks-3 &    heart-hungarian &                       bank &    lymphography &  pb-MATERIAL \\
   spectf &         arrhythmia &                      adult &     balloons &                    mushroom \\
  titanic &               bank &               steel-plates &             zoo &                   fertility \\
\bottomrule
\end{tabular}}
\end{center}
\label{table: uci dataset anecdotdal summary}
\end{table}
\end{small}

%

\subsection{Hyperparameter Optimization}
\label{section: hpo}
Hyperparameter optimization plays an important role in the machine learning community and can be the main factor in deciding whether a trained model performs at the state-of-the-art or simply moderate. The use of meta-features for this task has led to a significant improvement especially when used for warm-start initialization, the process of selecting initial hyperparameter configurations to fit a surrogate model based on the similarity between the target dataset and other available datasets, of Bayesian optimization techniques based on Gaussian processes~(\cite{DBLP:conf/icdm/WistubaSS15,DBLP:conf/aaai/LindauerH18}) or on neural networks~(\cite{DBLP:conf/nips/PerroneJSA18}). Surrogate transfer in sequential model-based optimization~(\cite{DBLP:journals/jgo/JonesSW98}) is also improved with the use of meta-features as seen in the state-of-the-art~(\cite{DBLP:journals/ml/WistubaSS18}) and similar approaches~(\cite{DBLP:conf/pkdd/WistubaSS16,DBLP:journals/corr/abs-1802-02219}). 
Unfortunately, existing meta-features, initially introduced to the hyperparameter optimization problem in~(\cite{DBLP:conf/icml/BardenetBKS13}), are engineered based on intuition and tuned through trial-and-error. We improve the state-of-the-art in warm-start initialization for hyperparameter optimization by replacing these engineered meta-features with our learned task-agnostic meta-features, proving further the capacity of the learned meta-features to correlate to unseen meta-tasks, (\textbf{D4}).
\subsubsection*{Baselines}
The use of meta-features has led to significant performance improvement in hyperparameter optimization. We follow the warm-start initialization technique presented by~(\cite{DBLP:conf/aaai/FeurerSH15}), where we select the top-performing configurations of the most similar datasets to the target dataset to initialize the surrogate model. By simply replacing the engineered meta-features with the meta-features from Dataset2Vec, we can effectively evaluate the capacity of our learned meta-features. Meta-features employed for the initialization of hyperparameter optimization techniques include a battery of summaries calculated as measures from information theory~(\cite{DBLP:conf/mdai/CastielloCF05}), general dataset features and statistical properties~(\cite{DBLP:journals/paa/ReifSGBD14}) which require completely labeled data. A brief overview of some of the meta-features used for warm-start initialization is summarized in Table~\ref{char}.  NS is not applicable in this scenario considering that hyperparameter optimization is done across datasets with different schema.
\begin{enumerate}
\item Random search~(\cite{bergstra2012random}): As the name suggests, random search simply selects random configurations at every trial, and has proved to outperform conventional grid-search methods.
\item Tree Parzen Estimator (TPE)~(\cite{DBLP:conf/nips/BergstraBBK11}), A tree-based approach that constructs a density estimate over good and bad instantiations
of each hyperparameter.
\item GP~(\cite{rasmussen2003gaussian}): The surrogate is modeled by a Gaussian process with a Mat\'ern 3/2 kernel
\item SMAC~(\cite{DBLP:conf/lion/HutterHL11}): Instead of a Gaussian process, the surrogate is modeled as a random-forest~(\cite{DBLP:journals/ml/Breiman01}) that yields uncertainty estimates\footnote{https://github.com/mlindauer/SMAC3.git}.
\item Bohamiann~(\cite{DBLP:conf/nips/SpringenbergKFH16}):  This approach relies on Bayesian Neural Networks to model the surrogate.
\end{enumerate}
On their own, the proposed baselines do not carry over information across datasets. However, by selecting the best performing configurations of the most similar datasets, we can leverage the meta-knowledge for better initialization. 

\subsubsection*{Evaluation Metrics}
We follow the evaluation metrics of~(\cite{DBLP:conf/pkdd/WistubaSS16}), particularly, the average distance to the minimum (ADTM). After $t$ trials, ADTM is defined as
\begin{equation}
 \text{ADTM}((\Lambda_t^D)_{D\in\mathcal{D}},\mathcal{D}) = \frac{1}{|\mathcal{D}|}\sum\limits_{D\in\mathcal{D}}
     \min_{\lambda\in\Lambda_t^D}\frac{y(D,\lambda)-y(D)^\text{min}}{y(D)^\text{max}-y(D)^\text{min}}
\end{equation}
with $\Lambda_t^D$ as the set of hyperparameters that have been selected by a
hyperparameter optimization method for data set $D := \in \mathcal{D}$ in the first $t$ trials
and $y(D)^{\min}$, 
$y(D)^{\max}$ the range of the loss function on the hyperparameter grid
$\Lambda$ under investigation.

\subsubsection*{UCI Surrogate Dataset}
To expedite hyperparameter optimization, it is essential to have a surrogate dataset where different hyperparameter configurations are evaluated beforehand. We create the surrogate dataset by training feed-forward neural networks with different configurations on 120 UCI classification datasets. As part of the neural network architecture, we define four layouts: $\square$ layout, the number of hidden neurons is fixed across all layers; $\lhd$ layout, the number of neurons in a layer is twice that of the previous layer; $\rhd$ layout, the number of neurons is half of the previous layer, $\diamond$ layout, the number of neurons per layer doubles per layer until the middle layer than is halved successively. We also use dropout~(\cite{DBLP:journals/jmlr/SrivastavaHKSS14}) and batch normalization~(\cite{DBLP:conf/icml/IoffeS15}) as regularization strategies, and stochastic gradient descent (SGD)~(\cite{bottou2010large}), ADAM~(\cite{DBLP:journals/corr/KingmaB14}) and RMSProp~(\cite{tieleman2012lecture}) as optimizers. SeLU~(\cite{DBLP:conf/nips/KlambauerUMH17}) represents the self-normalizing activation unit. We present the complete grid of configurations in Table~\ref{grid}, which results in 3456 configurations per dataset.

\begin{table}[h]
\caption{Hyperparameter configuration grid. We note that redundant configurations are removed, e.g. $\lhd$ layout with 1 layer is the same as $\square$ layout with 1 layer, etc.}
\centering
\begin{center}
\begin{tabular}{ll}
\toprule
Aspect & Values \\
\midrule
Activation   & ReLU, leakyReLU, SeLU \\
Neurons   & $2^n{}|{} n\in [2,4]$\\
Layers   & $1,3,5$\\
Layout & $\square$,$\lhd$,$\rhd$,$\diamond$\\
Optimizer & ADAM, SGD, RMSProp\\
Dropout&$0, 0.2, 0.5$\\
Batch Normalization& True, False\\
\bottomrule
\end{tabular}
\end{center}
\label{grid}
\end{table}



\subsubsection*{Results and Discussion}
\begin{figure*}[t!]
\centering
	{\includegraphics[width=1\columnwidth]{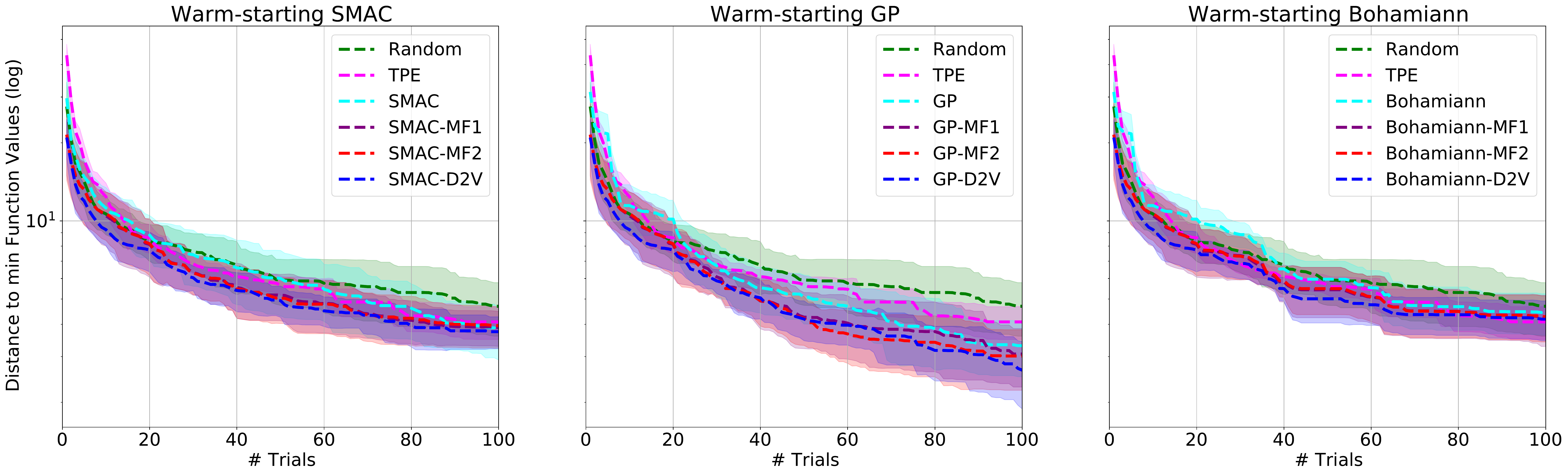}}
	\caption{Difference in validation error between hyperparameters obtained from warm-start initialization with different meta-features. By using our meta-features to warm-start these hyperparameter optimization methods, we are able to obtain better performance consistently. For all plots, lower is better.}
	\label{fig3}
\end{figure*}

The results depicted in Figure~\ref{fig3} are estimated using a leave-one-dataset-out cross-validation over the 5 splits of 120 datasets. We notice primarily the importance of meta-features for warm-start initialization, as using meta-knowledge results in outperforming the rest of the randomly initialized algorithms. By investigating the performance of the three initialization variants, we realize that with SMAC and Bohamiann, our learned meta-features consistently outperform the baselines with engineered meta-features. With GP, on the other hand, the use of our meta-features demonstrates an early advantage and better final performance. The reported results demonstrate that our learned meta-features, which are originally uncoupled from the meta-task of hyperparameter optimization prove useful and competitive, i.e. correlate (\textbf{D4}) to the meta-task. It is also worth mentioning that the learned meta-features, do not require access to the whole labeled datasets, making it more generalizable, Equation~\ref{mfe}.


\section{Conclusion}
We present a novel hierarchical set model for meta-feature learning based on the Kolmogorov-Arnold representation theorem, named Dataset2Vec. We parameterize the model as a feed-forward neural network that accommodates tabular datasets of a varying schema. To learn these meta-features, we design a novel dataset similarity learning task that enforces the proximity of meta-features extracted from similar datasets and increases the distance between meta-features extracted from dissimilar datasets. The learned meta-features can easily be used with unseen meta-tasks, e.g. the same meta-features can be used for hyper-parameter optimization and few-shot learning, which we leave for future work. It seems likely, that meta-features learned jointly with the meta-task at hand will turn out to focus on the characteristics relevant for the particular meta-task and thus provide even better meta-losses, a direction of further research worth investigating.

\section*{Acknowledgement}
This work is co-funded by the industry project \href{https://www.ismll.uni-hildesheim.de/projekte/ecosphere_en.html} {"IIP-Ecosphere: Next Level Ecosphere for Intelligent Industrial Production"}. Prof. Grabocka is also thankful to the Eva Mayr-Stihl Foundation for their generous research grant.

\bibliographystyle{spbasic}      
\bibliography{dataset2vec}
\end{document}